\documentclass[runningheads]{llncs}
\usepackage{graphicx}
\usepackage[utf8x]{inputenc} 
\usepackage{xcolor}
\usepackage{tabularx,ragged2e,booktabs,caption}
\newcolumntype{L}{>{\RaggedRight\arraybackslash}X}
\usepackage{caption}
\usepackage{subcaption}

\begin{document}
\title{Enriching Wikidata with Linked Open Data}
%
%

\author{Bohui Zhang
\and 
Filip Ilievski \and
Pedro Szekely}
\authorrunning{Zhang, Ilievski, and Szekely}
%
\institute{Information Sciences Institute, University of Southern California \\ \email{bohuizha@usc.edu, \{ilievski,pszekely\}@isi.edu}
}
%
\maketitle              
\begin{abstract}

Large public knowledge graphs, like Wikidata, contain billions of statements about tens of millions of entities, thus inspiring various use cases to exploit such knowledge graphs. However, practice shows that much of the relevant information that fits users' needs is still missing in Wikidata, while current linked open data (LOD) tools are not suitable to enrich large graphs like Wikidata. In this paper, we investigate the potential of enriching Wikidata with structured data sources from the LOD cloud. We present a novel workflow that includes gap detection, source selection, schema alignment, and semantic validation. We evaluate our enrichment method with two complementary LOD sources: a noisy source with broad coverage, DBpedia, and a manually curated source with a narrow focus on the art domain, Getty. Our experiments show that our workflow can enrich Wikidata with millions of novel statements from external LOD sources with high quality. Property alignment and data quality are key challenges, whereas entity alignment and source selection are well-supported by existing Wikidata mechanisms. We make our code and data available to support future work.

\keywords{Wikidata \and Linked Open Data \and Knowledge Enrichment \and Schema Alignment \and Semantic Validation}
\end{abstract}

\section{Introduction}
\label{sec:intro}

Commonly used knowledge graphs (KGs) today have wide coverage, recording information for many millions of entities through thousands of properties. Wikidata~\cite{vrandevcic2014wikidata}, the largest public KG, contains nearly 1.5B statements about 90M entities. This breadth of information inspires various use cases to exploit such KGs. Museum curators can use Wikidata to describe their art collections and artists comprehensively. Movie critics could quickly query and aggregate statistics about recent movies, and analyze them based on their genre or actor cast.

However, while Wikidata's data model allows for this information to be present, practice shows that much of the relevant information is still missing in Wikidata. For example, around half of the artists in Wikidata have a date of birth, and only 1.88\% of the movies recorded in 2020 have information about their cost. Thus, if a user wants to analyze the cost of the films produced in 2020, Wikidata will provide cost information for only 60 out of its 2,676 recorded films. 
As this information is unlikely to be found in an aggregated form, and gathering information about thousands of films is a tedious process, one must rely on automated tools that can enrich Wikidata with relevant information. Such information might be readily available in external linked open data (LOD)\footnote{\url{https://lod-cloud.net/}} sources like DBpedia~\cite{auer2007dbpedia} or LinkedMDB~\cite{hassanzadeh2009linked}; yet, no existing methods can enrich Wikidata with missing information. Conversely, link prediction~\cite{bordes2013translating} accuracy is not sufficient to use it to impute missing information with representation learning.

The traditional LOD workflow includes declarative language tools for schema mapping and templated rewritting of CSV files into RDF~\cite{das2011r2rml,knoblock2012semi,dimou2014rml}. A subsequent ontology alignment step~\cite{chu2020optimizing} can be employed to discover \textit{owl:sameAs} links between the two datasets. Yet, prior work~\cite{beek2018sameas} has shown that collapsing nodes based on \textit{owl:sameAs} relations is not feasible, as it leads to combining related but dissimilar nodes (e.g., Barack Obama and the US government) into a single node. Moreover, while prior efforts have focused on schema alignment between two sources, this challenge may be less significant when augmenting Wikidata, which provides 6.8k external identifier properties that explicitly link to entities in other sources.\footnote{\url{https://www.wikidata.org/wiki/Category:Properties\_with\_external-id-datatype}, accessed March 2, 2022.} Further challenge is data quality~\cite{piscopo2019we}, as external sources may contain information that is noisy, contradictory, or inconsistent with existing knowledge in Wikidata. Table~\ref{tab:dbpedia} presents candidate enrichment statements retrieved from DBpedia infoboxes. Here, only one out of four statements is correct, whereas the other three have an incorrect datatype, semantic type, or veracity. It is unclear how to employ principled solutions for schema alignment and quality validation to for large-scale and high-quality enrichment of Wikidata.

\begin{table}[!t]
\centering
\caption{Example statements found in DBpedia.} 
\label{tab:dbpedia}
\begin{center}
\begin{tabular}{ l | l | l }
\toprule
\bf DBpedia statement & \bf Wikidata mapping & \bf Assessment \\ \hline
\multicolumn{1}{m{5cm}|}{\texttt{dbr:Lesburlesque dbp:genre dbr:Burlesque}} & \texttt{Q6530279 P136 Q217117} & Correct \\ \hline
\multicolumn{1}{m{5cm}|}{\texttt{dbr:Amanda\_de\_Andrade dbp:position 'Left back'\@en}} & \multicolumn{1}{m{4cm}|}{\texttt{Q15401730 P413 'Left back'\@en}} & Wrong datatype \\ \hline
\multicolumn{1}{m{5cm}|}{\texttt{dbr:Diego\_Torres\_(singer) dbp:genre dbr:Flamenco}} & \texttt{Q704160 P2701 Q9764} & Wrong semantic type \\ \hline
\multicolumn{1}{m{5cm}|}{\texttt{dbr:Eternal\_Moment dbp:director dbr:Zhang\_Yibai}} & \texttt{Q5402674 P4608 Q8070394} & Logical, inaccurate \\
\bottomrule
\end{tabular}
\end{center}
\end{table}



In this paper, we investigate \textit{how to enrich Wikidata with freely available knowledge from the LOD cloud}. We introduce a novel LOD workflow to study the potential of the external identifier mechanism to facilitate vast high-quality extensions of Wikidata. Assuming that we can link Wikidata entities to external sources robustly through external identifiers, we start by aligning the entities automatically, followed by inferring the property in the external source that corresponds to the Wikidata property we seek to augment. This alignment of entities and properties yields candidate enrichment statements, which are rigorously checked through datatype comparison, semantic constraints, and custom validators. We demonstrate the effectiveness of the augmentation workflow on two LOD knowledge sources: Getty~\cite{harpring2010development} and DBpedia~\cite{auer2007dbpedia}. Getty is a manually curated domain-specific knowledge graph with a narrow focus on art, whereas DBpedia is a broad coverage KG which has been automatically extracted from Wikipedia. Extensive experiments on these sources show that our method facilitates vast and high-quality extension of Wikidata with missing knowledge.

We make our code and data available to facilitate future work.\footnote{\url{https://anonymous.4open.science/r/hunger-for-knowledge-315D/}}

\section{Related work}

Schema mapping languages, like R2RML~\cite{das2011r2rml}, Karma~\cite{knoblock2012semi}, and RML~\cite{dimou2014rml} enable users to map relational databases into RDF, optionally followed by a semi-automatic ontology alignment step. Recent ontology alignment methods~\cite{chu2020optimizing} typically align two ontologies in the vector space. Compared to prior works that align two schemas or ontologies, we seek to improve the coverage of a single large knowledge graph, by selective augmentation with external knowledge.

Ontology enrichment deals with noise, incompleteness, and inconsistencies of ontologies, by discovering association rules~\cite{d2016ontology}, or by extracting information from WWW documents~\cite{faatz2002ontology}. Ontology evolution seeks to maintain an ontology up to date with respect to the domain that it models, or the information requirements that it enables~\cite{zablith2015ontology}. Notably, there have been no efforts to enrich or improve the evolution of large-scale hyperrelational KGs, like Wikidata.

Prior work has attempted to enrich Wikidata with satellite data for a given domain, including frame semantics~\cite{mousselly2016enriching}, biodiversity data~\cite{waagmeester2019wikidata}, and cultural heritage data~\cite{faraj2019enriching}.  By learning how to represent entities in a KG like Wikidata, link prediction models~\cite{bordes2013translating} can predict missing values by associating them to known statements. Our work complements prior efforts that enrich Wikidata with domain-specific data or predict missing links, as we aim to devise a method for generic, large-scale enrichment of Wikidata with external LOD sources.

The linked open data cloud contains many identity links which could be explored to combine information from different sources. LOD Laundromat~\cite{beek2014lod} is a centralized infrastructure which captures a static version of the LOD cloud, which could be queried in order to obtain information from multiple sources at once. However, as we show in this paper, identity links between entities by themselves are insufficient for enrichment of knowledge graphs. The correspondence between the properties in two graphs may not be one-to-one, requiring inference about property paths. Furthermore, semantic validation needs to be performed to ensure that the mapped properties indeed have compatible semantics. In that sense, our work relates to prior work that explores identity links in the LOD cloud~\cite{beek2018sameas} or devises mechanisms to discover latent identity links~\cite{raad2017detection}. The goal of our work is different - to enrich Wikidata by finding high-quality knowledge in well-connected sources.

The automatic validation of our method relates to prior work on analyzing the quality of large KGs. Beek et al.~\cite{beek2018literally} propose a framework for analyzing the quality of literals on the LOD Laundromat, but it is unclear how to generalize this framework to entity nodes. Prior work has studied the quality of Wikidata~\cite{piscopo2017provenance,shenoy2021study} and compared it to the quality of other knowledge graphs, like YAGO~\cite{houcemeddine_turki_2020_4445363} and DBpedia~\cite{farber2018linked}. Compared to Wikidata, YAGO4  takes a step further and it defines constraints on domain and range, disjointness, functionality, and cardinality. The authors report that enforcing these constraints leads to a removal of 132M  statements from Wikidata, i.e., 28\% of all facts. In~\cite{shenoy2021study}, the authors apply five semantic constraints, including value type, on Wikidata in order to measure its quality. None of these works has investigated how to automatically validate external statements that can be used to enrich Wikidata. 

Wikidata's property constraint pages define existing property constraints and report number of violations for a single dump.\footnote{\url{https://www.wikidata.org/wiki/Help:Property\_constraints\_portal}}
Moreover, Wikidata includes several tools that monitor, analyze, and enforce aspects of quality, such as the primary sources tool (PST),\footnote{\url{https://www.wikidata.org/wiki/Wikidata:Primary\_sources\_tool\#References}} the Objective Revision Evaluation Service (ORES),\footnote{\url{https://www.wikidata.org/wiki/Wikidata:ORES}} 
and Recoin (``Relative Completeness Indicator'')~\cite{balaraman2018recoin}. Recoin computes relative completeness of entity information by comparing the available information for an entity against other similar entities. Our work complements these efforts by Wikidata, by providing mechanisms for automatic alignment and semantic validation of knowledge from the LOD cloud before it is submitted to Wikidata.


\section{Method}
\label{sec:method}

\begin{figure}[!t]
     \centering
         \includegraphics[width=1.0\textwidth]{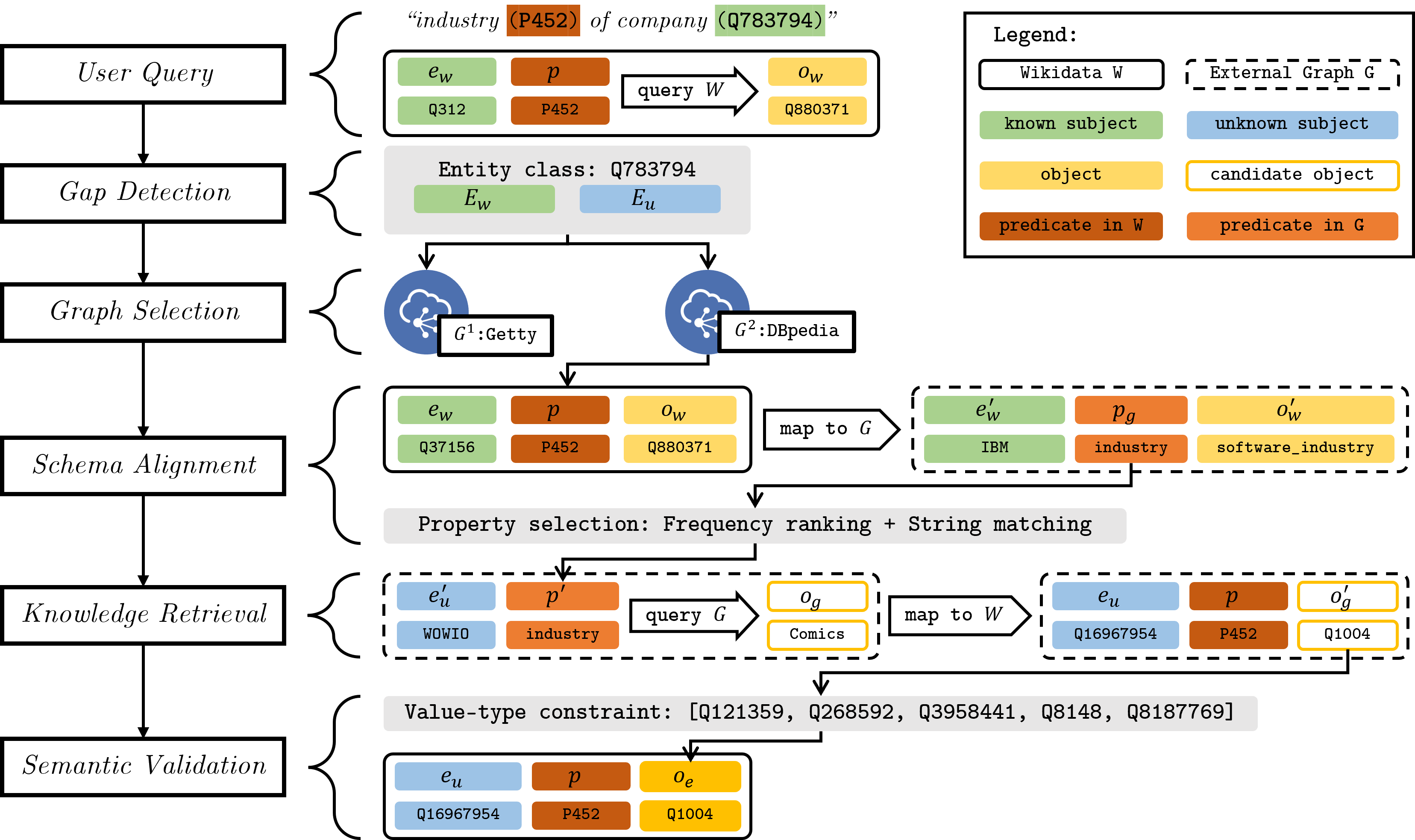}
         \caption{Our enrichment method, illustrated on enriching Wikidata with additional knowledge from DBpedia about the query ``industry of companies''.}
     \label{fig1}
\end{figure}

Our enrichment method is shown in Figure~\ref{fig1}. 
Given a user query, our method queries Wikidata ($W$), obtaining a set of known statements $S_w$ for subjects $E_w$, and a set of subjects with unknown values $E_u$. 
We call this step \textit{gap detection}, as it generates a set of entities for which we seek missing knowledge in order to satisfy the user's query needs. Considering that the LOD cloud contains other sources that are likely to contain information about the same entities, we perform a \textit{KG selection} step to determine a relevant set of KGs $G$ to consult for the entities in $E_u$. Here, the sources $G$ are assumed to overlap with $W$ in terms of the entities they describe.
Our \textit{schema alignment} step consolidates the entities and properties of $W$ with those of each $G$, since their entity and property identifiers are generally different. After aligning the two schemas, a \textit{knowledge retrieval} step queries the external KG $G$ to obtain new statements $S_g$ for the entities in $E_u$. A \textit{semantic validation} step is employed in order to ensure that the semantics of the newly found statements in $S_g$ corresponds to the semantics intended by Wikidata. The set of validated statements $S_e \subseteq S_g$ is finally used to enrich $W$. This procedure yields a more complete response to the user query consisting of a union of the original and the enriched statements, formally: $S_{total}=S_w \cup S_e$.

We show this procedure by querying Wikidata for ``industry of companies'' in Figure~\ref{fig1}. The gap detection step splits the overall set of target companies into two groups: 1) companies with known industry in Wikidata ($E_w$); and 2) companies with unknown industry ($E_u$), for which $W$ lacks industry information. The KG selection step decides that DBpedia may contribute missing knowledge for company industries. The schema alignment step relies on the statements in Wikidata, $S_w=\{(e_w, p, o_w) | e_w \in E_w\}$ about the known companies $E_w$. Each pair $(e_w, o_w)$ is aligned to a DBpedia-valued pair $({e_w}^\prime, {o_w}^\prime)$. Similarly, the unknown entities $e_u \in E_u$ are mapped in the same way to DBpedia entities ${e_u}^\prime$. Then, the external KG DBpedia is queried for property paths which correspond to the known subject-object pairs $({e_w}^\prime, {o_w}^\prime)$. The DBpedia path $p^\prime$ (\texttt{dbp:industry}) that corresponds to the Wikidata property $p$ is discovered with our method. 
The knowledge retrieval step obtains values from DBpedia, by querying for pairs $({e_u}^\prime, p^\prime)$ comprised of DBpedia entities and property paths, resulting in a new set of statements $S_g = \{({e_u}^\prime,p^\prime,o_g)\}$. 
The newly found values are inversely mapped back to `found' (candidate) Wikidata values ${o_g}^{\prime}$. Each of these newly obtained values is semantically validated, resulting in a subset of enrichment values ${o_e}$, which form the statements $(e_u,p,o_e)$.
\subsection{Gap detection}

We consider a structured query against a target knowledge graph $W$ for a query property $p$ (e.g., industry). 
The gap detection step generates a set of subject entities $E_w$ for which the value for the property $p$ is known in Wikidata, and a set of entities $E_u$ for which the value of the property $p$ is missing in $W$. $E_w$ and $E_u$ are subsets of the overall set of target entities and they are mutually disjoint, formally: $E_w \subseteq E$, $E_u \subseteq E$, $E=E_k \bigcup E_u$, and $E_w \bigcap E_u = \emptyset$.

This work focuses on finding values for entities in $E_u$ that have zero values for a property in Wikidata. We note that it is possible that the statements in $S_w$ do not fully answer the query for the entities $E_w$, as these entities may have multiple values for $p$, e.g., a politician may have several spouses throughout their life. Enriching multi-valued properties will be addressed in future work.



\subsection{Graph selection}

The graph selection step provides a set of LOD sources ${G}$ that can be used to enrich the results of the query.
In this work, we consider an automatically extracted general-domain KG (DBpedia) and a domain-specific curated KG (Getty). We experiment with using both KGs, or selecting one based on the posed query. 

\noindent \textbf{DBpedia}~\cite{auer2007dbpedia} is an open-source KG derived from Wikipedia through information extraction. Follow Wikipedia's guidelines, DBpedia's entity documents  are language-specific. DBpedia describes 38 million entities in 125 languages, while its English subset describes 4.58 millions of entities.
Large part of the content in DBpedia is based on Wikipedia's infoboxes: data structures containing a set of property–value pairs with information about its subject entity, whose purpose is to provide a summary of the information about that subject entity.
We use DBpedia infoboxes in order to enrich Wikidata, as this data is standardized, relevant, and expected to be extracted with relatively high accuracy.


\noindent \textbf{Getty}~\cite{harpring2010development} is a curated LOD resource with focus on art.
In total, Getty contains information about 324,506 people and 2,510,774 places. 
Getty consists of three structured vocabularies: (1) Art \& Architecture Thesaurus (AAT) includes terms, descriptions, and other information (like gender and nationality) for generic concepts related to art and architecture; (2) the Getty Thesaurus of Geographic Names (TGN) has 321M triples with names, descriptions, and other information for places important to art and architecture; and (3) the Union List of Artist Names (ULAN) describes names, biographies, and other information about artists and architects, with 64M statements. 
\subsection{Schema alignment}

As Getty and DBpedia have a different data model compared to Wikidata, we first align their schemas to Wikidata in order to query them based on missing information in Wikidata. The schema alignment consists of two sequential steps:
    
\noindent \textbf{1. Entity resolution} maps all known subject entities in Wikidata, $e_w$, unknown subjects $e_u$, and the known objects $o_w$ to external identifiers $e_w^\prime$, $e_u^\prime$, and $o_w^\prime$, respectively.
We map Wikidata nodes to nodes in external KGs automatically, by leveraging external identifiers and sitelinks available in Wikidata. In total, Wikidata contains 6.8K
external-id properties. 
Wikidata contains 46,595,392
sitelinks, out of which 5,461,631
link to the English Wikipedia pages. Our method leverages sitelinks to map Wikidata entities to DBpedia nodes,\footnote{Wikipedia page URIs can trivially be translated to DBpedia URIs.} while for Getty we use vocabulary-specific external-id properties in Wikidata: Art \& Architecture Thesaurus ID (\texttt{P1014}) for AAT items, Getty Thesaurus of Geographic Names ID (\texttt{P1667}) for TGN items, and Union List of Artist Names ID (\texttt{P245}) for ULAN items. We note that, while the entity mapping is automatic, the selection of the external-id property itself in the current method is manual, e.g., the user has to specify that P1667 should be used for TGN identifiers.

\noindent \textbf{2. Property alignment} We map the property $p$ from Wikidata to a corresponding property path $p^\prime$ in $G$ by combining structural and content information. We query $G$ for property paths $p_g$ with maximum length $L$ that connect the mapped known pairs ($e_w^\prime$, $o_w^\prime$). We aggregate the obtained results, by counting the number of results for each property path $p_g$ that connects the known pairs. We preserve the top-10 most common property paths, and we use string similarity to select the optimal one. Here, we rerank the top-10 candidates based on Gestalt Pattern Matching~\cite{gestalt}.\footnote{Our empirical study showed that this function leads to comparable accuracy like Levenshtein distance, but it is more efficient.} If the most similar property path has similarity above a threshold (0.9), then we select it as a mapped property $p^{\prime}$, otherwise we select the top-1 most frequent property.
In the example in Figure~\ref{fig1}, the Wikidata property \texttt{P452} (industry) would map to \texttt{dbp:industry} in DBpedia, which is both the top-1 most frequent property in the aligned results, and it has maximum string similarity with $p$. We expect that string similarity and value frequency can complement each other. For example, the top-1 most frequent property for \texttt{P149} (architectural style) is \texttt{dbp:architecture}, while string similarity correctly reranks the top-10 candidates to select the right mapping \texttt{dbp:architecturalStyle}. The property chain $p^\prime$ can be quite complex, e.g., Wikidata's \texttt{P19} (place of birth) maps to a 4-hop path in Getty: \texttt{foaf:focus} $\rightarrow$ \texttt{gvp:biographyPreferred} $\rightarrow$ \texttt{schema:birthPlace} $\rightarrow$ \texttt{skos:exactMatch}.



\subsection{Knowledge Retrieval}

The schema alignment step produces external identifiers for the unknown entities $e_u^\prime$ and an external property path $p^\prime$ that corresponds to the property in the original query. The user can query the external graph $G$ with the mapped subject-property pair $(e_u^\prime, p^\prime)$, in order to automatically retrieve knowledge. In Fig.~\ref{fig1}, an example of $(e_u^\prime, p^\prime)$ pair is (\texttt{dbr:WOWIO, dbp:industry}). We denote the candidate objects found with this step with ${o_g}$. As the candidate object identifiers belong to the external graph, the user can perform inverse entity resolution by following sitelinks or external identifiers from the external graph $G$ to $W$. This step results in newly found Wikidata objects $o_g^{\prime}$ for the unknown entities $e_u$, completing their statements $(e_u, p, o_g^{\prime})$.


\subsection{Semantic Validation}

Despite the schema alignment, the candidate objects ${o_g^{\prime}}$ may be noisy: they may have a wrong datatype (e.g., date instead of a URI), an incorrect semantic type (e.g., a nationality instead of a country), or a literal value that is out of range (e.g., death year 2022). We trim noisy objects with three validation functions.


\noindent \textbf{Datatypes} Each Wikidata property has a designated datatype and the candidate objects ${o_f}$ have to conform with it. For instance, the spouse values are expected to be Qnodes, movie costs should be numeric values with units (e.g., 4 million dollars), while founding years -- dates with a year precision (e.g., 2015). To infer the expected datatype of a property, we count the datatypes of the known objects $o_w$, and select the top-1 most common datatype. This function returns a subset of statements $S_{v1}$ with candidate objects that belong to the expected datatype.

\begin{figure}[!t]
\centering
\includegraphics[width=0.48\textwidth]{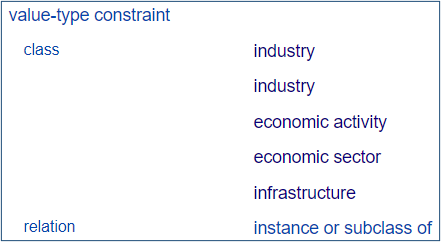}
\caption{Example value-type constraint of the property industry (\texttt{P452}). The values associated with this property should belong to one of the following types: \texttt{[industry, industry, economic activity, economic sector, infrastructure]}, whose respective Qnodes are \texttt{[Q8148, Q268592, Q8187769, Q3958441, Q121359]}. The type can be either encoded as an instance-of (\texttt{P31}) or a subclass-of (\texttt{P279}) property. There are no entities in Wikidata which are exceptions for which this property constraint.} 
\label{fig3}
\end{figure}

\noindent \textbf{Property constraints} We validate the object values ${o_f}$ further based on property constraints defined in Wikidata. Specifically, we use value-type constraints to validate the semantic type of the objects in $o_n$. Value type constraints are similar to property range constraints~\cite{shenoy2021study}, but they provide a more extensive definition that includes exception nodes and specifies whether the type property is: \texttt{P31} (instance of), \texttt{P279} (subclass of), or both. Figure \ref{fig3} shows an example of a value type constraint for the property \texttt{P452} (industry). 
We automatically validate the value type of all statements for a property, by comparing their object value to the expected type. 
Following~\cite{shenoy2021study}, we encode the value type constraint for a property as a KGTK~\cite{ilievski2020kgtk} query template. 
Each template is instantiated once per property
, allowing for efficient constraint validation in parallel. Constraint violations for a property are computed in a two-step manner: we first obtain the set of statements that satisfy the constraint for a property, and then we subtract this set from the overall number of statements for that property. The constraint validation function yields a set of validated statements $S_{v2}$.

\noindent \textbf{Literal range validation} We validate date properties (e.g., date of birth) by ensuring that they do not belong to the future, i.e., that every recorded date is smaller than the year 2022. This function outputs a set of valid statements $S_{v3}$.


The set of validated statements is the intersection of the results returned from the three validation functions: $S_v=S_{v1} \cap S_{v2}$ for Qnodes and $S_{v1} \cap S_{v2} \cap S_{v3}$ for date values. This validated statements in $S_e$ have the form $(e_u, p, o_e)$. The total set of statements for the user query becomes $S_{total}=S_w \bigcup S_e$.
\section{Experimental Setup}
\label{sec:experiment}

\noindent \textbf{Knowledge graphs} We experiment with batch enrichment for Wikidata properties that have value-type constraints, 955 in total. We use the Wikidata 2021-02-15 dump in a JSON format and the sitelinks file from 2021-10-27. We use the 2021-12-01 DBpedia infobox file in a Turtle (.ttl) format.\footnote{\url{https://databus.dbpedia.org/dbpedia/generic/infobox-properties/}} We select its cannonicalized version, as it ensures that its subjects and objects can be mapped to English Wikipedia pages. 
For the Getty vocabularies, we download the current dump in N-Triples (.nt) from their website.\footnote{\url{http://vocab.getty.edu/}}

\noindent \textbf{Evaluation} To evaluate the quality of the overall enrichment, we randomly sample candidate statements and annotate their validity manually. Specifically, we sample 100 statements from the DBpedia set and 30 statements from Getty. Two annotators annotate 130 statements independently by searching each one on Internet and determine whether it is correct or not. The annotation resulted in nearly perfect agreement, and the remaining two conflicts were resolved with a discussion between the annotators. After this annotation, we observe that 20/100 of the DBpedia statements and 11/30 of the Getty statements are correct, while the rest are incorrect. We annotate three reasons for incorrect statements: wrong datatype, wrong semantic type, and inaccurate information. Out of the 80 incorrect DBpedia statements, 66 have incorrect datatype, 7 incorrect semantic type, and 7 are inaccurate. For Getty, 11 are correct, 0 have incorrect datatype, 17 incorrect semantic type, and 2 are inaccurate.

To investigate the quality of our property alignment, we use property mappings provided by \texttt{owl:equivalentProperty} in DBpedia and \texttt{P1628} (equivalent property) in Wikidata as ground truth to evaluate the property alignment of our method. We dub this data \texttt{Equivalence}. In the property mapping pairs we collected, each Wikidata property is mapped to one or multiple DBpedia properties. In total, 88 Wikidata properties are mapped to 101 DBpedia properties.
We formulate a task where the goal is to map a Wikidata property to its corresponding DBpedia property/properties. We accept any of the DBpedia mappings according to the gold standard data as correct. Besides correct and incorrect mappings, we annotate an intermediate category of close match for properties that match partially. We show two F1-values of our method: hard, which only counts exact matches, and soft, which includes partial matches. We compare our method to three baselines. The \textit{string matching} and \textit{frequency matching} baselines are ablations of our method that only consider string similarity or frequency, but not both. The third baseline embeds all DBpedia labels with BERT~\cite{bert} and uses cosine distance to select the closest property label.






\noindent \textbf{Implementation}
We implement our method using the Knowledge Graph ToolKit (KGTK)~\cite{ilievski2020kgtk}. 
For Getty we obtain paths with maximum length of $L=4$, for DBpedia $L=1$. In the schema alignment step, we count property frequency based on a sample up to 200,000 pairs of known subjects and objects $(e_w, o_w)$. 

\section{Results}
\label{sec:results}

In this Section, we present six experimental findings that concern the potential of our method, its overall accuracy, its accuracy per component, and its consistency when the external information is covered by Wikidata.

\begin{table}[!t]
\centering
\caption{Batch enrichment results when using DBpedia, Getty, and both KGs. $|S_*|$ shows numbers of statements. In total, we consider 955 properties. $|p^\prime|$ shows the numbers of properties mapped to each of the KGs.}
\label{tab:batch}
\begin{center}
\begin{tabular}{c | r | r r r r }
\toprule
 & $|p^\prime|$ & $|S_w|$ & $|S_g|$ & $|S_e|$ & $|S_{total}|$ \\ 
\midrule
\bf DBpedia & 582 & 106,104,551 & 41,309,864 & 21,023,187 & 127,127,738 \\ 
\bf Getty & 3 & 195,153 & 10,518 & 5,766 & 200,919 \\ 
\bf Both & 582 & 106,104,551 & 41,320,382 & 21,028,953 & 127,328,657 \\ 
\bottomrule
\end{tabular}
\end{center}
\end{table}

\noindent \textbf{Finding 1: Our method can enrich Wikidata with millions of statements about millions of entities.}
Table~\ref{tab:batch} shows the results of our method for all properties in Wikidata that have a value-type constraint. Out of 955 Wikidata properties, our method is able to align 582 properties with DBpedia and 3 can be mapped with Getty. For the Wikidata properties aligned with DBpedia and Getty, we gather 21 million statements, enriching the original Wikidata knowledge by 16.54\%. Interestingly, while the original Wikidata focuses on a broad coverage of entities, our method is able to enrich more properties for a smaller set of entities, signifying a higher density and a more narrow focus. The $|S_g|$ column shows that our method collects 41 million candidate statements from DBpedia and Getty, out of which 21 million pass the semantic validation.
The median number of novel statements per property is 982.
For 161 (27.66\%) properties, our method provides double or more statements relative to the original set of statements. The relative increase of knowledge is the lowest for the properties \texttt{P538} (fracturing), \texttt{P209} (highest judicial authority), \texttt{P1283} (filmography), and \texttt{P534} (streak color). Meanwhile, the properties \texttt{P66} (ancestral home), \texttt{P500} (exclave of), and \texttt{P3179} (territory overlaps) are relatively sparse in Wikidata, and receive many more statements from DBpedia. 
Comparing the two external KGs, we observe that DBpedia overall contributes many more statements and entities than Getty. DBpedia is also bringing a higher enrichment per property, averaging at 16.54\% vs 2.87\% for Getty. 

\noindent \textbf{Finding 2: The overall quality of the enriched statements is relatively high.}
Table~\ref{tab:annotation} shows that our method can distinguish between correct and incorrect statements with a relatively high accuracy of over 88\%. As the majority of the candidates are incorrect, we observe that the F1-score is lower than the accuracy. The precision of our method is lower than the recall on both DBpedia and Getty, which indicates that most of the disagreement of our method with human annotators is because of false positives, i.e., incorrect statements identified as correct by our semantic validator. This indicates that our semantic validation is accurate but it is not complete and it can benefit from additional validators. This observation is further supported by the relatively higher precision and recall of our method on Getty in comparison to DBpedia. As Getty is manually curated and enforces stricter semantics, is has a smaller range of data quality aspects to address, most of which are already covered by our method. The quality issues in the case of DBpedia are heterogeneous, as a result of its automatic extraction and lack of curation.  
We find that out of 130 triples, 52 had incorrect property mappings, and the semantic validation is able to correct 44 of them. For example, Wikidata property \texttt{P208} (executive body) got mapped to \texttt{dbp:leaderTitle} in DBpedia, and its value \texttt{Q30185} was not allowed by the Wikidata constraint for \texttt{P208}. We evaluate the property alignment and the semantic validation in more detail later in this Section.


\begin{table}[!t]
\centering
\caption{Evaluation results on randomly sampled 130 candidate triples: 100 from DBpedia and 30 from Getty. We show the accuracy, precision, recall and F1-score of our semantic validation on this subset. Getty does not have values with a wrong datatype (`-').}
\label{tab:annotation}
\begin{center}
\begin{tabular}{c | r | r r r | r r r r }
\toprule
KG & Accuracy & Precision & Recall & F1-score & \multicolumn{4}{c}{Accuracy per category} \\
& & & & & Correct & Datatype & Sem. type & Inaccurate \\
\midrule
\bf DBpedia & 87.00\% & 61.54\% & 84.21\% & 71.11\% & 80.00\% & 96.97\% & 71.43\% & 28.57\% \\
\bf Getty & 93.33\% & 84.62\% & 100.0\% & 91.67\% & 100.0\% & - & 100\% & 0\% \\ 
\bf Both & 88.46\% & 69.23\% & 90.00\% & 78.26\% & 87.10\% & 96.97\% & 91.67\% & 22.22\% \\
\bottomrule
\end{tabular}
\end{center}
\end{table}

\noindent \textbf{Finding 3: Property mapping performance is relatively high, but sparse properties are difficult.}
Table~\ref{tab:batch} showed that around 40\% of the target Wikidata properties had no match found in DBpedia with our method. To investigate whether this is because of misalignment between the two schemas or a limitation of our method, we evaluate the property alignment precision and recall of our method on the \texttt{Equivalence} data. The results are shown in Table~\ref{tab:prop_map}. Our method achieves the best F1-score for both the soft match cases (89\%) and the hard match cases (66\%).\footnote{We also manually evaluate the property matching methods on a separate randomly chosen set of 20 properties, and observe similar results.}
As frequency and string matching are ablations from our method, their lower performance supports our decision to combined them to get the best of both worlds. For instance, frequency matching tends to prefer more general properties over specific ones, mapping \texttt{P30} (continent) to \texttt{dbp:location}. Thanks to string matching, our method predicts the right property \texttt{dbp:continent} in this case. Conversely, string matching is easily confused by cases where the labels are close but the actual meaning is not related, e.g., it maps \texttt{P161} (cast member) to \texttt{dbp:pastMember}, whereas our method correctly maps it to the ground truth result \texttt{dbp:starring} owing to its frequency component. 
As our method still largely relies on the frequency of statements, we hypothesize that its performance decreases for properties with fewer example statements. 
To investigate this hypothesis, we divide the ground truth properties into four quartiles (of 22 properties) based on the descending size of their original Wikidata statements. We evaluate the accuracy or our property matching per quartile. We note that the performance of the first three quartiles with larger number of statements is relatively better than the last quartile, which indicates that the precision of our method is positively correlated with the size of known statements. This limitation can be addressed in the future with more robust methods, e.g., based on learned property representations.

\begin{table}[!t]
\centering
\caption{Evaluation results on known aligned properties between DBpedia and Wikidata. We compare against exact match and language model baselines. We also show the performance of our method per property quartile, where the quartiles are based on the number of examples for a property.}
\label{tab:prop_map}
\begin{center}
\begin{tabular}{ c | r r r | r r r }
\toprule
Method & \multicolumn{3}{c}{Hard match} & \multicolumn{3}{c}{Soft match} \\
 & Precision & Recall & F1-score & Precision & Recall & F1-score \\
\midrule
BERT Embedding & 47.73\% & 47.73\% & 47.73\% & 72.73\% & 72.73\% & 72.73\% \\ 
Frequency Matching & 52.33\% & 51.14\% & 51.72\% & 81.40\% & 79.55\% & 80.46\% \\
String Matching & 52.27\% & 52.27\% & 52.27\% & 86.36\% & 86.36\% & 86.36\% \\ 
Our Method & \textbf{66.28\%} & \textbf{64.77\%} & \textbf{65.52\%} & \textbf{89.53\%} & \textbf{87.50\%} & \textbf{88.51\%} \\ 
\midrule
Our method (Q1) & 71.43\% & 68.18\% & 69.77\% & 90.48\% & 86.36\% & 88.37\% \\
Our method (Q2) & 63.64\% & 63.64\% & 63.64\% & 100.0\% & 100.0\% & 100.0\% \\
Our method (Q3) & 81.82\% & 81.82\% & 81.82\% & 86.36\% & 86.36\% & 86.36\% \\
Our method (Q4) & 47.62\% & 45.45\% & 46.51\% & 80.95\% & 77.27\% & 79.07\% \\
\bottomrule
\end{tabular}
\end{center}
\end{table}

\noindent \textbf{Finding 4: Semantic validation can detect wrong datatypes and semantic types.}
Table~\ref{tab:batch} shows that the semantic validation has a very large impact on the results: out of 41 million candidate statements initially found by our method, around half of them satisfy the datatype and semantic type constraints of our validation. 
The compatibility ratios are similar for both Getty and DBpedia (50.89\% to 54.82\%), which is surprising, considering that DBpedia has been largely extracted automatically and is error-prone, whereas Getty is well-curated and considered an authority.
To study the precision and recall of our semantic validation, we annotate three reasons for incorrect statements: wrong datatype, wrong semantic type, and inaccurate information. We found that (Table~\ref{tab:batch}) our method performs well on identifying correct statements (F1-score 93.10\%), as well as detecting errors due to wrong datatypes (F1-score 96.97\%) and incorrect semantic types (F1-score 91.67\%). 
Our method performs relatively worse when the statements satisfy the value-type constraints but are inaccurate. For example, the enriched statement (\texttt{Q6712846 P19 Q49218}) for the property \texttt{P19} (place of birth) from Getty is logical but inaccurate, since the value \texttt{Q49218} satisfies the value-type constraints while it is not the actual birth place of \texttt{Q6712846}. Among 130 triples, our method produces 7 false positive cases that are factually incorrect. These results are expected, given that our method is designed to filter out illogical information, while analyzing veracity is beyond its current scope.




\begin{table}[!t]
\centering
\caption{Evaluation results for data consistency. Numbers of Wikidata statements, enriched statements, overlapping entity-property values, agreeing statements, and disagreeing statements are counted. The agreement ratio $r_{agree}$ is calculated by $|S_{agree}| / |S_{overlap}|$.}
\label{tab:agree}
\begin{center}
\begin{tabular}{ c | r r r r r r }
\toprule
KG (Property) & $|S_w|$ & $|S_e|$ & $|S_{overlap}|$ & $|S_{agree}|$ & $|S_{disagree}|$ & $r_{agree}$ \\
\midrule
DBpedia (P19) & 2,711,621 & 467,976 & 884,078 & 461,089 & 422,989 & 52.15\% \\
DBpedia (P20) & 1,080,900 & 119,161 & 219,447 & 128,523 &  90,924 & 58.57\% \\
Getty (P19) & 65,411 & 2,939 & 16,304 & 13,607 & 2,697 & 83.46\% \\	
Getty (P20) & 50,295 & 2,556 & 14,722 & 12,594 & 2,128 & 85.55\% \\
\bottomrule
\end{tabular}
\end{center}
\end{table}

\noindent \textbf{Finding 5: Most results for functional properties are consistent between external graphs and Wikidata, many disagreements are due to different granularities.}
The analysis so far focused on the statements that are novel, i.e., provide novel object values for subject-property pairs that do not have them in Wikidata. Here, we measure consistency between the validated statements from the external graph with the known statements of Wikidata, we select two functional properties: \texttt{P19} (place of birth) and \texttt{P20} (place of death), and count the agreements and disagreements for both DBpedia and Getty. The results in Table~\ref{tab:agree} for \texttt{P19} and \texttt{P20} show that 52-59\% of the overlapping statements between Wikidata and DBpedia, and 83-86\% of the overlapping statements between Wikidata and Getty coincide. Qualitative inspection of the disagreements reveals that many of the disagreements are due to different granularity choices between the two graphs. For instance, in Wikidata, the place of death of \texttt{Q1161576} (Daniel Lindtmayer II) is \texttt{Q30978} (Central Switzerland) which is a region, while Getty provides the specific city \texttt{Q4191} (Lucerne).


\begin{figure}[!t]
     \centering
     \begin{subfigure}[b]{0.49\textwidth}
         \centering
         \includegraphics[width=\textwidth]{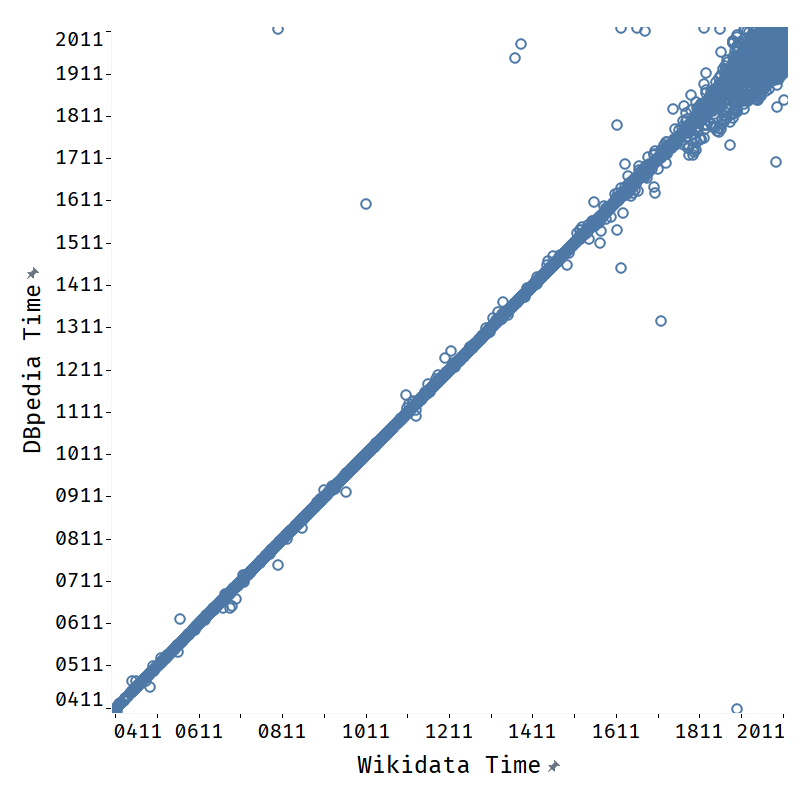}
         \caption{DBpedia}
         \label{a}
     \end{subfigure}
     \hfill
     \begin{subfigure}[b]{0.49\textwidth}
         \centering
         \includegraphics[width=\textwidth]{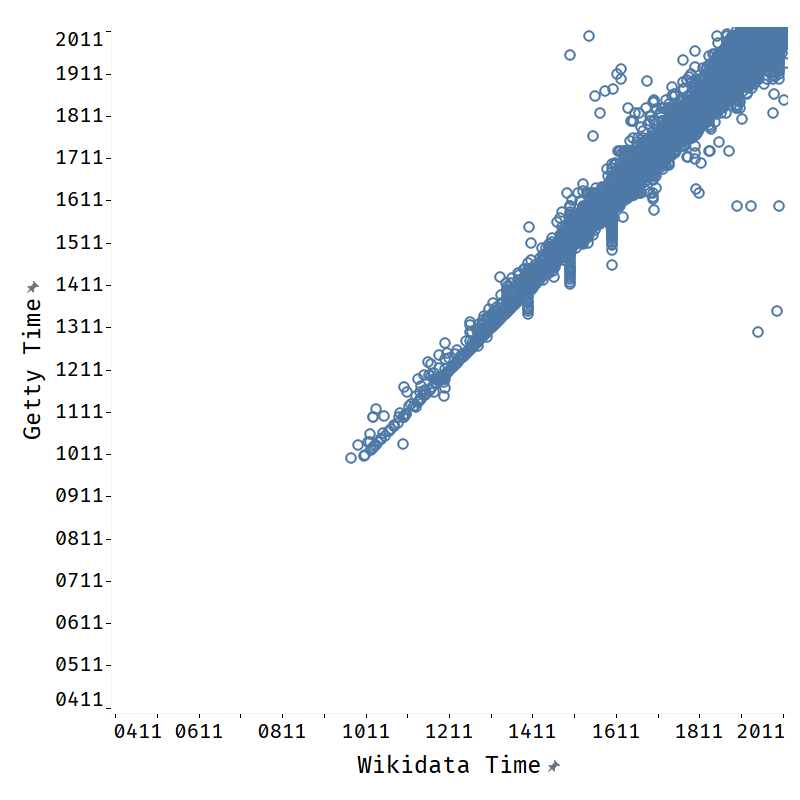}
         \caption{Getty}
         \label{b}
     \end{subfigure}
     \caption{Scatter plots of literal consistency for \texttt{P570} (date of death) between Wikidata and the external graph: DBpedia (a) and Getty (b). The plots show the subject-property pairs for which both Wikidata and the external KG have a value. Ticks on both the X- and the Y-axis represent years.}
     \label{fig:literals}
\end{figure}

\noindent \textbf{Finding 6: The method can help enrich literals as well.}
Our method can also be used to enrich literal information about entities. We run our method on two functional properties: \texttt{P569} (date of birth) and \texttt{P570} (date of death). We compare the obtained results from the external graphs to those found in Wikidata, for entity-property pairs where both Wikidata and the external graph have a value. We compare the results on the finest granularity provided by the the two graphs, which is dates for DBpedia and years to Getty. 
The results for property \texttt{P570} in Figure~\ref{fig:literals} show a clear trend of the points in scatter plots distributed along the line of $y=x$, which shows high consistency of the date data between the Wikidata and the external KGs.\footnote{We observe a similar trend for \texttt{P569}.} Specifically, we observe that the agreement rate with Wikidata values is 89.28\% (1,271,862 out of 1,424,526) for DBpedia and 82.39\% for Getty (125,913 out of 152,824).
From Getty and DBpedia, our method can enhance Wikidata with novel \texttt{P569} values for 35,459 entities and novel \texttt{P570} values for 20,664 entities.

\section{Discussion and Future Work}
\label{sec:discussion}

Our enrichment method has been shown to quickly retrieve millions of novel property values in the LOD cloud for entities in Wikidata. 
As some of the LOD knowledge is extracted in an automatic way, ensuring quality is important - our semantic validation based on datatypes and constraints found around half of the candidate statements to be invalid. 
Analysis of a subset of the enriched statements revealed that the accuracy of our method is close to 90\%, which is reasonably high.
Still, our method is merely a step towards the ambitious goal of addressing the notorious challenge of sparsity of today's large KGs~\cite{dong2014knowledge}. Here, we discuss three key areas of improvement for our method:

\noindent \textbf{1. Evaluation on more KGs} - We showed that our method is effective with two external KGs: a general-domain and automatically extracted knowledge graph, DBpedia, and a domain-specific, well-curated knowledge graph, Getty. As Wikidata is still largely incomplete after this enrichment, we can use the 6.8k external identifier properties provided by Wikidata to enrich with thousands other sources. While we expect that our method can be applied on these thousands of sources, an in-depth investigation of the potential and the quality of this enrichment is beyond the scope of the current paper. 


\noindent \textbf{2. Semantic validation} - Our method validates candidate statements through datatype and value type constraints. Value type constraints ensure semantic type compatibility of the retrieved statements, yet they are only one of the 30 property constraint types defined in Wikidata. Other Qnode constraints in Wikidata can be employed to generalize our method to properties that do not have a value type constraint. Qnode-valued statements can be further validated via constraints like \texttt{one-of (Q21510859)}, whereas literals can, for instance, be checked with \texttt{range (Q21510860)}.



\noindent \textbf{3. Validating veracity} Table~\ref{tab:annotation} shows that our method performs relatively well on detecting statements with incorrect datatype or semantic type, whereas it is usually unable to detect inaccurate statements. The accuracy of detecting inaccurate statements ranges from 0 to 28\%. As discussed by Piscopo~\cite{piscopo2019we}, veracity is a key aspect of quality of knowledge in KGs. Our method can be further enhanced with models that detect KG vandalism~\cite{heindorf2016vandalism} or estimate trust of sources (e.g., through references)~\cite{piscopo2017provenance} to estimate veracity.



\section{Conclusions}

Recognizing the notorious sparsity of modern knowledge graphs, such as Wikidata, and the promise of linked data information, like external identifiers, to facilitate enrichment, this work proposed a method for enriching Wikidata with external KGs found in the LOD cloud. The method consisted of five steps: gap detection, external graph selection, schema alignment, knowledge retrieval, and semantic validation. We evaluated our method on enriching Wikidata with two LOD graphs: DBpedia and Getty. Our experiments showed that our LOD-based method can enrich Wikidata with millions of new high-quality statements in a short time. High-quality enrichment is achieved based on large-scale automated semantic validation based on datatypes and value type constraints, as well as a hybrid algorithm for property alignment. A key future direction is evaluating the generalization of our method on thousands of LOD sources that Wikidata points to, which opens novel challenges of source selection, more extensive semantic validation, and trust. We make our code and data available to facilitate future research on LOD-based knowledge enrichment.

\bibliographystyle{splncs04}
\bibliography{references}
%




\end{document}